\documentclass[twoside,11pt]{article}

\usepackage[hyphens]{url}
\usepackage{jmlr2e}
\usepackage{enumitem}
\usepackage{times,mathtools}
\usepackage{color}
\usepackage{options}
\usepackage{packages}
\usepackage{macros}
\usepackage{microtype,xparse,tcolorbox}
\usepackage{graphicx}
\usepackage{footnote}
\usepackage{ragged2e}
\graphicspath{ {./Images/} }

\newcommand{\fcal}{\mathcal{F}}
\newcommand{\xcal}{\mathcal{X}}

\allowdisplaybreaks

\begin{document}

\thispagestyle{plain}
\title{A Comprehensive Survey on Machine Learning Techniques and User Authentication Approaches for Credit Card Fraud Detection}


\newcommand{\HRule}{\rule{\linewidth}{0.5mm}} 
\center 
 



\HRule \\[0.4cm]
{\LARGE \bfseries A Comprehensive Survey on Machine Learning Techniques and User Authentication Approaches for Credit Card Fraud Detection} 
\HRule \\[1 cm]


 {Niloofar Yousefi, Marie Alaghband, Ivan Garibay\\
  Department of Industrial Engineering and Management Systems\\
  University of Central Florida\\
  Orlando, Florida, USA \\
  \texttt{niloofar.yousefi@ucf.edu, marieh.alaghband@knights.ucf.edu, igaribay@ucf.edu}\\[1 cm]
  }


\center
\justify{
\begin{abstract}
With the increase of credit card usage, the volume of credit card misuse also has significantly
increased, which may cause appreciable financial losses for both credit card holders and financial
organizations issuing credit cards. As a result, financial organizations are working hard on
developing and deploying credit card fraud detection methods, in order to adapt to ever-evolving,
increasingly sophisticated defrauding strategies and identifying illicit transactions as quickly as
possible to protect themselves and their customers. Compounding on the complex nature of such
adverse strategies, credit card fraudulent activities are rare events compared to the number of
legitimate transactions. Hence, the challenge to develop fraud detection that are accurate and
efficient is substantially intensified and, as a consequence, credit card fraud detection has lately become a
very active area of research. In this work, we provide a survey of current techniques most relevant
to the problem of credit card fraud detection. We carry out our survey in two main parts. In the first part, we focus on studies utilizing classical machine learning models, which mostly employ traditional transnational features to make fraud predictions. These models typically rely on some static physical characteristics, such as \emph{what} the user knows (knowledge-based method), or \emph{what} he/she has access to (object-based method). In the second part of our survey, we review more advanced techniques of user authentication, which use behavioral biometrics  to identify an individual based on his/her unique behavior while he/she is interacting with his/her electronic devices. These approaches rely on \emph{how} people behave (instead of \emph{what} they do), which cannot be easily forged. By providing an overview of current approaches and the results reported in the literature, this survey aims to drive the future research agenda for the community in order to develop more accurate, reliable and scalable models of credit card fraud detection.
\end{abstract}
}


\begin{flushleft}
\justify{
\section{Introduction}
Credit card fraud is perpetrated in many different shapes and forms, but can be broadly categorized into the following: `application', `electronic or manual card imprint', `mail non-receipt', `lost or stolen', `counterfeit' and `card not present' fraud. Application fraud generally happens when other people apply for and obtain new credit cards using false personal information. This typically arises in conjunction with identify theft, as fraudsters usually steal supporting document needed to substantiate fraudulent applications. Electronic or manual card imprint fraud is usually committed through skimming information on the magnetic strip of the card and then using it to perform a fraudulent transaction. Mail non-receipt is when credit cards in the post are intercepted by fraudsters before they reach the card holder. Another type of fraud can happen though lost or stolen cards, in which the credit card is taken from the card holder's possession, either through theft or misplacement. The criminals then use the card to make payments. Counterfeit card fraud is done by skimming a fake magnetic swipe card, which holds all the details of the card. An exact copy of the card then will be created using the fake strip.

The most common form of credit card fraud is when the card is not present (i.e. phone, mail, internet transactions). In this type of fraud, the fraudulent transaction does not involve the presentation of a tangible card, and it is performed remotely, through phone, mail, or internet, where retailers are unable to physically check the card or the identity of the card holder. In such type of fraud, the details of the credit card are obtained without the card holder's knowledge, either by a skimming process (\eg\ when employees use an unauthorized ‘swiper’ that downloads the encoded information), or from receipts thrown away by the customer. One of the most striking factors about this form of fraud is that fraudsters may be able to even assume the actual identity of the victim. This enables fraudsters to obtain access to the victim's bank account and ask the bank for a change of address so that all correspondence from the bank will go to a fake address. This increases the time that fraudsters have to fraudulently make charges on their victims' credit cards. This form of fraud occurs through hacking online databases, which allow fraudsters to take control of peoples' bank accounts, and let assume false identities. By cold calling and pretending to be from the bank or the police, fraudsters may obtain complete information of their victims needed to purchase goods or services in the name of the card holder. 


\subsection{Preliminaries}
\subsubsection{Machine Learning}
With an ever-increasing amount of data, it is almost impossible for a human programmer or specialist to detect a meaningful pattern in data and translate it for the future use. For this reason, the application of machine learning has widely spread throughout computer science domains, where information extraction from large data sets is required. These applications include but are not limited to spam filtering, web searching, ad placement, face and voice recognition, recommender systems, credit scoring, drug design, fraud detection, and stock trading. In this section, we provide some background on the main concepts underlying machine learning.

Learning covers a wide range of processes, which are difficult to define precisely. Consequently, machine learning has waded into several branches, each of which deals with a different type of learning task. However, the shared features of all machine learning models is that they automate the process of an inductive inference including, observe a phenomenon, build a model based on the observed phenomenon, and make predictions using the constructed model. Formally, there are several types of learning frameworks such as supervised, semi-supervised, unsupervised, reinforcement, transduction, and learning to learn. The two most widely adopted machine learning methods are supervised learning, which trains the algorithm on predefined labeled datasets, and unsupervised learning, which provides the algorithm unlabeled training data to allow learning the patterns and relationships in the input data. In the following, we explore these two learning frameworks in more detail.

\subsubsection{Supervised Learning}
Supervised learning is commonly used to use historical data to predict statistically likely future events. In this framework, the phenomenon is defined as some instance-label pairs. The learning  model is then constructed as a mapping function from the instances to the labels. For example, consider an input space $\xcal$ as the set of objects we want to label. Also, assume that the output space $\mathcal{Y}$ denotes the set of possible labels. We then assume that the training data $\left\lbrace (X_1, Y_1), . . . ,(X_n, Y_n)\right\rbrace $ are identically and independently drawn from an unknown distribution $P$ defined on $\xcal \times \mathcal{Y}$. Now, given the training data, the objective of a learning algorithm is to choose a function $f :\xcal \rightarrow \mathcal{Y}$ among the functions in the hypothesis class $\fcal$. In other words, the objective of a learning algorithm is to choose a function $f$ from the hypothesis space $\fcal$ that minimizes the empirical risk, and quantify the consistency of function $f$ with the training data as 
\begin{align}
\label{ERM}
R_{n}(f) := \frac{1}{n} \sum_{i=1}^{n} \boldsymbol{1}_{f(X_{i})\neq Y_{i}}
\end{align}

The learning algorithm, which is called Empirical Risk Minimization, is based on the idea of choosing a predictor function $f \in \fcal$ which minimizes \eqref{ERM}. However, in practice, a regularizer is usually imposed on $\fcal$ to prevent overfitting. This leads to regularized empirical risk minimization algorithms, which solves the following problem:
\begin{align}
\min_{f \in \fcal}  \frac{1}{n} \sum_{i=1}^{n} \boldsymbol{1}_{f(X_{i})\neq Y_{i}} + \lambda \Omega(f)
\end{align}
where $\Omega(f)$ is the  regularization function, and $\lambda$ is the parameter associated to  $\Omega(f)$.

Regression and classification are the two most common supervised approaches. Recommendation and time series prediction are two popular examples of problems built on top of classification and regression, respectively.

\subsubsection{Unsupervised Learning}
Unlike supervised learning, in unsupervised frameworks, we only observe the instances $\{X_{1},\ldots,X_{n}\}$, which are not associated with any labels. The goal of an unsupervised learning algorithm is then to model the underlying structures, patterns, similarities, and differences in the data that can be used for decision making or predictions about future inputs.

A big challenge associated to unsupervised learning is that it's almost impossible to come up with a reasonable objective measure of the algorithm's accuracy, since there is no gold standard (like a target output or label). Therefore, choosing an appropriate measure to asses the similarity of the instances or to compare different structures in the data is always challenging in this learning framework.
 
Dimensionality reduction and clustering  are two classic examples of unsupervised learning. Dimensionality reduction creates/extracts a smaller number of variables (features) from the original variables (features) that describe the data sufficiently. Since dimensionality reduction is not relevant or applicable to our study here, we will not elaborate further on these methods. 
 
Clustering is about partitioning of data into groups (clusters) such that the objects in each group share some common characteristics among each other. A popular clustering model that minimizes the clustering error is known as the $K$-means algorithm.  K-means clustering aims to partition the given data into $K$ clusters with minimal variation within each cluster, and maximal variation across the clusters. The idea is to initialize the $K$ cluster by randomly selecting $K$ points as cluster centers. The algorithm then assigns a data point to their nearest center and updates the clustering centers by calculating the average of the members belonging to the clusters. This process repeats the relocating-and-updating steps until the algorithm converges. The convergence criteria is typically set based on a predefined number of iterations and/or the difference on the value of the distortion function. Besides the challenge of selecting an appropriate similarity/dissimilarity measure, other limitations of clustering approaches in general are 1) choosing the right number of clusters along with the initial centers of the clusters, as these two initial starting conditions heavily affect the performance of the algorithm; 2) numerical data is required as the $K$-means clustering algorithm works only with numerical data, limiting its application to real world problems.

\section{Literature Review}
We carried out our survey in two main parts. In the first part, we focused on studies addressing the problem of credit card fraud detection using classical \ac{ML} models. These approaches typically employ traditional transnational features (such as type of the card, type of the transaction, place of the transaction, etc.) to make fraud predictions. All these methods are based on user authentication or identification models, which rely on some static physical characteristics, such as what the user knows (knowledge-based method), or what he/she has access to (object-based method). All these knowledge or object based authentication approaches suffer from an obvious disadvantage:
tokens can be stolen, misplaced or guessed by fraudsters. Therefore, this type of models is not capable of distinguishing an authorized user from an impostor, who fraudulently acquire the token or knowledge of the authorized person. To achieve a more satisfactory level of security, more advanced techniques have been proposed for user authentication/identification, which use behavioral biometrics to identify an individual based on his/her unique behavior while he/she is interacting with his/her electronic devices. The main advantage of these approaches is that the security relies on \emph{how} people behave (instead of \emph{what} they do), which cannot be easily forged. 

In the second part of our survey, we review some of the recent studies addressing the problem of user authentication using behavioral biometrics. It is worth pointing out that since there have not been enough studies that use behavioral biometrics for fraud detection, we focused our review on general user authentication approaches employing behavioral biometrics to identify authorized users from impostors. We would also like to mention that with regard to the second part of the survey, we were mainly concerned about the different authentication schemes used in different works; meaning that we reviewed and compared studies in terms of different behavioral features, rather than approaches that they used for user authentication.
  
\subsection{Fraud Detection Approaches Based on Traditional \ac{ML} Models}
Searched through ``Google Scholar" for recent scientific papers published in the context of credit card fraud detection using \ac{ML} models. ``Credit Card Fraud Detection" and ``Machine Learning" were the most useful keywords to find%
relevant articles. In this stage, by reviewing the abstract and title of the collected papers, we identified articles, which use \ac{ML} approaches to address the problem of credit card fraud detection. In the next stage, we considered and reviewed only recently published or highly-cited papers among all collected ones. 

This section presents a review of the \ac{ML} approaches adopted for credit card fraud detection. According to literature, both supervised and unsupervised algorithms can be used for credit card fraud detection. In the following, we will present an overview of relevant studies using these two approaches to address the problem at hand. 

\subsubsection{Supervised \ac{ML} Algorithms for Credit Card Fraud Detection} 
Assuming the availability of annotated transactional data, supervised techniques can be used to discover patterns associated to genuine and fraudulent classes of transactions. The main advantage of these methods is that they have low false positive (normal transactions considered as frauds) rates, since the model can learn existing fraud signatures and patterns in the available data. However, one shortcoming of the supervised approach is that the model's learning (classier) is based on limited available fraud records; therefore, it might not be successful in predicting novel fraudulent behaviors that may not have been seen in the past. This might lead to a high false negative (fraudulent transactions considered legitimate) rate in the fraud detention system. These methods also need to deal with the challenge of unbalanced class sizes, as genuine transactions generally far outnumber the
fraudulent ones. 

Many studies exploited the strength of supervised \ac{ML} algorithms to predict fraudulent credit card activities. \textbf{\ac{LR}} is a broadly utilized technique that has been heavily used in initial fraud detection studies. Although these methods are well-understood, easy to implement, and have a well established history with fraud detection, they have limited power when dealing with non-linear data, which makes them unsuitable for complex fraud detection problems. For example, the authors in \citep{jha2012employing} use a \ac {LR} to show the improved performance of a transaction aggregation strategy in creating suitable derived attributes, which help detect credit card fraud. Using real-life data of transactions from an international credit card operation, they evaluate the performance of their proposed model. The results show the importance of transaction types, product types, and/or merchant types as discriminator features to identify credit card fraud. Another study in \citep{sahin2011detecting} compares the performance of \ac{LR} and \ac{ANN} in credit card fraud detection with a real data set. The empirical results of this study illustrated the advantage of \ac{ANN} over \ac{LR} on the test data, and it shows an equal performance of these two models over training data. Another interesting observation in this work is that the \ac{LR} model overfits the training data, and its overfitting behavior becomes even more remarkable as the number of training data increases. Also, \citep{rushin2017horse}'s research compares the predictive power of logistic regression against gradient boosted trees and deep learning. Using a real dataset---containing $80$ million account-level transactions with $69$ attributes collected in an eight-month time period---this study demonstrates the advantage of deep learning as the most accurate model in fraud prediction. Also, it indicates that \ac{LR} has the worst performance among all models, which could be due to its inability in finding hidden relationships.

\textbf{\ac{ANN}}s are other popular methods that have been applied for supervised credit card fraud detection. Neural networks are highly adaptive for complex data structures, however, they require high computational power for training, and they are prone to local minima, overfitting, and noise.  One study in \citep{patidar2011credit} addresses the problem of credit card fraud detection using neural networks. In their study, a genetic algorithm is applied to design the network and make
decisions about the network topology, number of hidden layers, and number of nodes. The authors in \citep{kim2002neural} propose an aggregated framework based on neural networks and a fraud density map. In this approach, a neural classifier generates a fraud ratio from the feature vector, which is then combined with a a fraud density generated from a fraud density map. The transactional data from $1977$ from a credit company in Korea has been used to evaluate this model, and the results indicate the high effectiveness of this model compared to the plain neural network classifier. Another study \citep{wang2018privacy} developed a privacy-preserving distributed deep neural network algorithm, which enables banks and other entities to share their data without revealing sensitive information. Using a real-world fraud detection data set containing multi-million transactions, the study in \citep{wang2018privacy} achieves performance (measured by AUC) comparable to the non-private baseline.

\textbf{\ac{SVM}} is another popular method that has been widely applied to detect frauds in the credit card industry. \ac{SVM}s have been proven to be successful in a variety of classification tasks, including fraud detection. The strength of SVMs comes from two unique features: first, their capability to work in high dimensional feature spaces without any additional computational complexity, achieved by using kernel functions to map the data from their original input space to a high dimensional feature space, wherein the instances are more likely to be linearly separable. This property of \ac{SVM}s enables non-linear classification problems like the ones arising in fraud detection to be solved. Another distinctive attribute of \ac{SVM}s is the way they achieve global solutions with good generalization performance. This property is due to the fact that the optimization problem in this framework minimizes an upper bound of the generalization error. This property makes \ac{SVM} more robust in its prediction task. In one study \citep{chen2005personalized}, the performance of \ac{SVM} has been compared against \ac{ANN} in investigating the time-varying fraud problem. The results indicate that \ac{SVM} and \ac{ANN} are comparable in training. However, \ac{ANN} tends to overfit training data, and therefore shows worse performance for future data prediction. Another study in \citep{lu2011research} developed a detection model based on class weighted \ac{SVM}, which is expected to more effectively handle the imbalance nature of the credit card transnational data. Using a real dataset from a Chinese bank, it is illustrated that their proposed model is effective in solving the credit card fraud detection problem leading to higher precision compared to the 
C-SVM, Back Propagation Neural Network (BPNN), Decision Tree, and Naive Bayes classifier. Moreover, with the application of a real data set, the authors in \citep{sahin2011detecting} compare the performance of SVM and Decision Tree in credit card fraud detection.The results of this study show that with a large number of training samples the performance of \ac{SVM}-based models is comparable to that of generated from decision tree approaches. Although, the Classification and Regression Trees have a higher true positives (fraudulent) rate. Another study in \citep{chen2006new} developed a \ac{SVM}-based model, which is shown to be effective in predicting the positive (fraudulent) samples more accurately. Their proposed approach has been shown to outperform two other approaches, namely the over-sampling (replicating the data in the minority class) and the adding-sampling (adding the data in the minority class), which both are commonly used to moderate the problem of imbalance in the data set. Another work \citep{dheepa2012behavior} also shows the effectiveness of \ac{SVM}in credit card fraud detection. The study in \citep{dheepa2012behavior} proposes a \ac{SVM} model, which uses the spending behavior of users to detect fraudulent transactions. Principle Component Analysis has been employed for feature selection and it has been shown to be effective in achieving lower false alarm rate.

\textbf{\ac{DT}} is another type of classifier which has been adopted by researchers to build fraud detection models. These methods are easy to implement, display, and understand with a low computational power requirement. Despite the flexibility and interpretability of \ac{DT}s, they can be unstable and highly sensitive to skewed class distributions. Also, the selection of splitting criteria can significantly affect the tree's classification performance. The study in \citep{shen2007application} compares the prediction accuracy of \ac{ANN}, \ac{DT}, and \ac{LR} in credit card fraud detection. Using a real-world data set, this work demonstrates the advantages of neural networks and logistic regression over decision tree in solving the problem under investigation. Using real credit card data from a bank, the authors in \cite{sahin2013cost} evaluate the performance of their cost-sensitive \ac{DT}-based approach in identifying fraudulent credit card transactions. The results show that the proposed model outperforms the traditional models in fixed-cost \ac{DT}, \ac{ANN}, and \ac{SVM} not only in terms of accuracy and true positive rate, but also with respect to a newly defined performance metric---suitable for the credit card fraud detection problem---which is the Saved Loss Rate (SLR). This is the saved percentage of the potential financial loss calculated with the available usable limits of the cards from which fraudulent transactions are committed. In a similar study in \citep{bahnsen2015example}, an example-dependent cost-sensitive decision tree algorithm was developed. The effectiveness of this model was examined using three real-world data sets including one provided by a large European card processing company. Using a newly defined cost-based impurity measure, they compared the performance of their proposed model against classical \ac{DT}, \ac{LR}, and Random Forest algorithms. The results indicate that the model outperforms the alternatives in terms of saving. Also, it has been shown that compared to classical \ac{DT} algorithm, their algorithm leads to a smaller tree with less complexity, which is easier to interpret and analyze.

The instability and sensitivity of \ac{DT}s have been addressed by adopting ensemble methods to create a forest of random trees. \textbf{\ac{RF}} is an ensemble of decision trees, which is more robust to overfitting and noise in the data. Also, they are computationally efficient, as each tree is generated independently. However, the performance of these models highly depend on the strength of each tree as well as the correlation between trees. Overfitting can easily occur when using these type of models. Also, \ac{RF}s have limited interpretability power due to the multitude of decision trees that
make up the ensemble. Several studies have evaluated the performance of \ac {RF} against other \ac{ML} approaches. For example, using real-world data of transactions from an international credit card operation, the study in \citep{bhattacharyya2011data} shows the performance advantages of \ac{RF} in capturing more fraud cases, with fewer false positives. Also, the authors in \citep{dal2015credit} have used and compared several \ac{RF}-based models to address the problem of concept drift (\ie\
customers' habits evolve). Their experiments on two real-world credit card data sets indicate that alert precision can be substantially improved by their proposed approach. Another study in \citep{dal2014learned} shows the superiority of \ac{RF} over \ac{NN} and \ac{SVM} on a real credit card dataset provided by a payment service provider in Belgium. Using a real credit card data set and several performance measures such as AP, AUC, and  PrecisonRank, they aim to address the challenging problem of learning with unbalanced data in credit card detection. In another work in \citep{van2015apate}, the authors compare the predictive power of \ac{RF} against \ac{LR} and \ac{NN}. Their experiments on a data set with more than three million transactions indicate that \ac{RF} outperforms its competitors in terms of accuracy and AUC. Moreover, \ac{RF} were found to outperform \ac{SVM}, \ac{LR} and \ac{KNN} in another comparative study in \citep{whitrow2009transaction}. Using two real-life data sets and a cost performance measure, they show that \ac{RF} gives the best results on both data sets, especially with aggregated data. They experiments also indicate that transaction aggregation technique has a major impact on the performance of classifiers for fraud detection.

\textbf{\ac{NB}} classifiers are other commonly used fraud detection techniques, which use probabilistic classifiers based on Bayes conditional probability to classify each sample into the class that it is most likely to belong to. They are easy to interpret and effective, especially with high dimensional input data. Moreover, they allow the integration of expert knowledge to uncertain statements. However, the predictive power of these models is highly affected by the assumption of conditional independency among features in the data, which leads to reduced accuracy in the presence of redundant attributes. A study in \citep{mohammed2018scalable} investigates the the suitability of several \ac{ML} algorithms including \ac{NB} classifier to detect credit card fraud with highly imbalanced massive data sets. With the help of two real world data sets, it has been show in \citep{mohammed2018scalable} that \ac{NB} technique is comparatively faster than the RF and Balanced Bagging Ensemble (BBE) classifier in detecting fraud. It should be noted that the precision rate is too low, which leads to many false alarms. Another research paper in \citep{mahmud2016evaluation} investigates the performance of various machine learning algorithms in detecting credit card fraud. Some measurements such as classification accuracy and fraud detection rate were used to evaluate different models. The results show that \ac{DT}-based models outperform \ac{NB} algorithm in terms of classification accuracy. Also, in order to build a cost sensitive detection system, a Bayes minimum risk approach was proposed in \cite{bahnsen2013cost}, which takes into account the real financial costs of credit card fraud detection. Using a real transactional data set, this study shows that the proposed framework is capable of reducing the cost compared to state-of-the-art techniques such as \ac{LR}, \ac{DT} C4.5 and \ac{RF}. In another study in \citep{mahmoudi2015detecting}, the authors evaluate the performance of linear Fisher discriminant analysis against \ac{NB}, \ac{ANN}, and \ac{DT}. Using a real-world datad set taken from an anonymous bank in Turkey, they show that their proposed approach outperforms alternatives based on not only on classical performance measures but also saved total available limit. 

\textbf{\ac{KNN}} algorithms have been successfully used in credit card fraud detection. Based on this technique, a new instance query is classified based on its $K$ nearest neighbors. The effectiveness of this model is influenced by the distance metric used to locate the nearest neighbors, and also the parameter $K$ which determines the number of neighbors needed to classify the new sample. Therefore, these algorithms are extremely sensitive to noise. In a comparative study in \citep{yeh2009comparisons}, using real data from a bank (a cash and credit card issuer) in Taiwan, the authors compared the predictive performance of several data mining methods including \ac{KNN}, \ac{ANN}, \ac{DT}, and \ac{NB} classifiers, as well as \ac{LR}, and discriminant analysis. Based on their results, \ac{KNN} classifiers have the lowest error rate. However, they show that \ac{KNN} does not perform better than the \ac{NB} classifier, \ac{ANN}, and \ac{DT}, if 
as the performance measure. Also, the work in \citep{pun2012improving} proposes a meta classifier based on \ac{KNN}, \ac{NB}, and \ac{DT} to address the problem of fraud detection. Using $11$ months of transactional data collected from a Canadian bank, the proposed algorithm shows significant improvement over \ac{NN}-based methods used by the bank to score fraudulent transactions. Also, the savings improvement evaluation was employed to assess the predictive performance of algorithms in fraud detection. Using UCSD Data Mining Contest 2009 Dataset (anonymous and imbalanced), the authors in \citep{seeja2014fraudminer} compared the performance of their proposed frequent pattern mining algorithm (based on frequent itemset mining) against \ac{KNN} and a few other \ac{ML} techniques. For this comparison, they used 4 classification metrics: fraud detection rate, false alarm rate, balanced classification rate, and Matthews correlation coefficient. The results of this study show the performance improvement of the proposed model over \ac{KNN}, \ac{NB}, \ac{SVM}, and \ac{RF} in terms of all 4 metrics. Also, it has been shown that \ac{KNN} performs better than \ac{SVM} on fraud detection rate (sensitivity), and it outperforms \ac{RF} and \ac{SVM} in terms of balanced classification rate. Interestingly, the results of the study in \citep{seeja2014fraudminer} indicate that \ac{KNN} is very competitive with the proposed frequent pattern mining approach in terms of false alarm rate.

\subsubsection{Unsupervised \ac{ML} Algorithms for Credit Card Fraud Detection}
The second type of approach deals with unsupervised techniques, which detect changes in behavior or unusual transactions. In these approaches, the legitimate user behavioral model is learned, then activities with enough departure from the norm are detected as frauds. An advantage of unsupervised methods is that they are more powerful than supervised approaches in detecting previously unseen types of frauds. Also, since they do not need labeled data, these methods can be useful in applications where no prior knowledge is available. 

One of the most known unsupervised algorithms used in fraud detection is clustering. For example, \textbf{\ac{KM}} is a simple and efficient clustering method that partitions unlabeled samples into $K$ disjoint clusters such that the square of distance between the points and centroid of that cluster is minimized. Although \ac{KM} clustering algorithms are simple and easy to implement, they are very sensitive to the initial cluster centers, which are randomly selected. This makes \ac{KM} algorithms vulnerable to outliers, which are especially relevant in fraud detection contexts. Moreover, similar to \ac{KNN} methods, in \ac{KM}s the parameter $K$ should be chosen appropriately, which requires input from domain experts. If not, it bears the burden of additional computation to find the optimal value of $K$. Another disadvantage of the clustering methods, including \ac{KM}s, is the difficulty of choosing an appropriate metric to measure the distance between observations. For example, it is not an easy task to combine categorical and numerical attributes in a good clustering metric, since samples may group differently on some subsets of attributes than they do on others. This may lead to instability problems in clustering-based methods. A K-means clustering algorithm is used in \citep{Vaishali2014} to detect fraudulent transactions. They generated an artificial data set including  transaction ID, transaction amount, transaction country, transaction date, credit card number, merchant category ID, and cluster ID. $4$ clusters were used to group credit card transactions into low, high, risky, and high risk. The results of their experiments showed that in most cases, the fraudulent activities could be correctly identified, although there were a few cases wherein a non-fraudulent activity was incorrectly detected as fraud. Also, a combination of Hidden Markov Model (HMM) and K-Means algorithms was used in \citep{Kumari2017} to identify the fraudulent activities on credit cards. In their proposed framework, a K-means clustering algorithm is first applied on the historical data to categorize customers based on their spending behavior in terms of having high, medium, and low transactions. Then the HMM produces an  output in the form of the probability of a transaction being fraudulent. In another study in \citep{chang2014analysis}, an X-means algorithm (a variant of K-means) was developed to cluster fraudsters as Aggressive, Classical, Luxury, or Low-profile. A fraud detection method was then performed to classify suspects into legitimate users or fraudsters in different classes, which was showed it could lead to improved overall detection accuracy. Also, the authors in \citep{behera2015credit} used a \ac{KM}-based clustering approach to address the problem of credit card fraud detection. Using the clustering algorithm, the transactions were grouped based on the spending patterns of the cardholders. A transaction is considered suspicious if its distance to the center of the cluster exceeds a pre-set threshold. The suspicious transactions are kept for further analysis and classification using a feed forward \ac{NN}. The statistical analysis of the results on a simulated data set yielded up to 93.90\% True Positive and less than 6.10\% False Positive. The study in \citep{jiang2018credit} used a \ac{KM} clustering method to cluster cardholders to three different groups:
low, medium and high transaction. Then, they utilized a window-sliding strategy to aggregate the transactions in each group, and then extracted a collection of specific behavioral patterns for each cardholder based on this. At the end, they applied a set of classifiers to detect fraudulent transactions in each group. The results of their experiments on simulated data showed improvement over \ac{LR}- and \ac{RF}-based models. 

\textbf{\ac{SOM}} is an unsupervised neural network learning model, which has been used to form customer profiles and visualize fraudulent patterns. In \ac{SOM}, the transaction data is grouped into genuine and fraudulent sets through the process of self-organization, which is an iterative tuning in the weights of neurons in the network. A new sample is then
fed into \ac{SOM}; if it is similar to all previous instances from a
genuine set, it is considered legitimate, or if it is similar to past examples of fraudulent sets, it is classified as fraud. \ac{SOM}s are very efficient and can handle large and high dimensional datasets, due to its visualization facilities. However, the lack of a real objective function makes it difficult to compare the solution of various \ac{SOM} models against each other. Besides that, like other neural network based models, choosing the optimal size of the \ac{SOM} requires expert knowledge and/or extensive computational evaluations. A fraud detection method based on \ac{SOM} is proposed in \citep{olszewski2014fraud}. In the first step of their algorithm, a \ac{SOM} visualization was performed on the multidimensional data of the user accounts. Then, a threshold-type binary classification algorithm was applied to detect fraudulent accounts. Their experimental study on a real data set demonstrates the benefits of data visualization, which transforms the input high-dimensional information into a 2-dimensional image, which is more interpretable even by non-experts. The study in \citep{agaskar2017unsupervised} proposed a fraud detection model which detected fraudulent transactions using records of the amount and location details of previous transactions carried out by customers. After obtaining the clusters from the SOM algorithm, they suggested re-validating the clusters using association rules on each cluster. Another \ac{SOM}-based fraud detection model was proposed in \citep{deng2009combining}, wherein a \ac{KM} clustering algorithm is applied on the results of SOM to avoid unclear clustering boundaries of nodes of SOM. Also, 100 financial statements of Chinese listed companies were used as experimental samples.

\subsection{User Authentication Approaches}
Designing secure and reliable user authentication systems has become an important task to protect users’ private information and data. Thus, it is relevant and articles studying this topic in the literature should be discussed. In order to do so, we did a through research on the recent articles from ``Google Scholar" and "science direct" resources. In this phase, we identified relevant studies published recently for deeper analysis. This selection was done based on if the research paper is pursuing a novel, interesting, or relevant approach which can be extended to the case of credit card fraud detection. Other recent surveys in user authentication task can be addressed by \cite{meng2015surveying}, \cite{kunda2018survey}, \cite{barkadehi2018authentication}, and \cite{neal2016surveying}.

\subsubsection{Password-Based Authentication}\label{subsec:Password-Based Authentication}
Password-based authentication approaches such as the Personal Identification Number (PIN) and Graphical passwords (\cite{dunphy2010closer}) are broadly used for user authentication. A recent study by \cite{meng2016multiple} on recalling multiple password interference on touch screen patterns and text passwords shows that with three accounts, users in the unlock pattern condition can perform better than users in text password condition.

\paragraph{}However, several articles discussed that these user authentication schemes are usable and convenient but highly insecure (\cite{de2012touch}). This is because users are more likely to choose simple passwords due to long-term memory limitations. As a result, passwords are usually easy to guess and remember (\cite{yan2004password} and \cite{florencio2007large}). Additionally, authentication credentials can be easily stolen via shoulder surfing (\cite{tari2006comparison} and \cite{kim2010multi}), in which hackers can use direct observation techniques to infer users’ data. Finally, graphical passwords can be recovered through side channel attacks (\cite{aviv2010smudge}), i.e., Android unlock patterns can be identified via a smudge attack, where attackers can extract recently touched locations by inspecting
smudges.

\paragraph{}Due to the aforementioned issues of password-based authentication, research has discussed using a new approach based on the measurements of human actions called biometric-based authentication. The following subsection addresses biometric-based authentication methods. 

\subsubsection{Biometrics-Based Authentication}
Here, we will discuss biometric-based authentication approaches for user authentication, which can be broadly classified into the two categories: Physiological Authentication and Behavioral Authentication schemes.

\paragraph{} \textbf{Physiological Authentication}: Approaches that use measurements from the human body and its physical characteristics, such as a fingerprint (\cite{maio2002fvc2000},
\cite{numabe2009finger}), face (\cite{wallace2012cross}, \cite{cardinaux2006user}, \cite{wallace2011inter}), voice, iris/retina (\cite{pillai2016robust}, \cite{mansour2016iris}) and hand/palm (\cite{zhou2009crease}, \cite{huang2008palmprint}) to classify users, are known as physiological authentication approaches. \cite{gunson2011usability} discusses voice recognition authentication in detail, in which they address the problem of using customer voices and comparing sentences versus digits for authentication in an automated telephone banking system.
They compared voice print authentication types in an experiment consisting of $204$ telephone banking customers, and found that using voice print authentication based on digits is more convenient and makes a significant impact in customer acceptance. \cite{dai2011multifeature} proposed a novel recognition algorithm using several features such as density, orientation, and principal lines for high resolution palmprints and designed an estimation algorithm. They performed their algorithm on a database containing $14,576$ full palmprints and discovered that density is very useful for palmprint recognition.
\paragraph{}Despite the uniqueness of physiological biometrics, which enables reliable authentication system construction, they usually require additional and special hardware to scan and/or recognize physiological features. Another limitation of this approach it allows access during for the whole session after the initial verification (\cite{meng2012touch}). This may provide the opportunity for an impostor to gain
access to a session in progress
and retrieve sensitive information. Additionally, environmental factors (such as different viewing angles, poor illumination, and background noises) can diminish the accuracy and reliability of physiological based authentication schemes (\cite{zheng2014you} and \cite{phillips2011introduction}). In addition, some of the features such as iris and retina scanning are very expensive (\cite{meng2018enhancing}). 
\paragraph{}Due to these limitations of physiological approaches, this part of our survey is geared towards investigating the studies that employ behavioral biometrics for user authentication purposes. It is noteworthy that some researchers consider only physical characteristics as part of biometric authentication schemes (such as \cite{kunda2018survey}), while in this survey, we consider biometric based authentication schemes as including both physiological and behavioral biometrics. 

\paragraph{} \textbf{Behavioral Authentication}: The second type of biometric approache, known as behavioral authentication,  uses human actions to authenticate users. It is shown that there is a significant difference between user's behavior while interacting touch screen surfaces. In a study by (\cite{sharma2017user}) on mobile devices it is shown that "\textit{As long as the user interface for a mobile application remains consistent, user behavior while interacting with the user interface also remains consistent}". As a result, several studies in the literature address behavioral authentication schemes, which may use different features to authenticate users and in general can be categorized as one of the following:
\begin{itemize} 
\item Keystroke Dynamics (Based on typing characteristics of the user)

\paragraph{}Keystroke dynamics are considered one of the most important features of behavioral biometrics (\cite{jiang2007keystroke}) due to its dependence on the time and the typing skill of the user (\cite{chuda2009multifactor}). As a result, several studies have been published in this area. As one of the early researches in the area, \cite{bergadano2002user} describe a new biometric measure of the typing characteristics of users that controls the instability of keystroke dynamics. They test their method on a data set of 154 individuals and achieved an $4\%$ average False Alarm Rate. \cite{clarke2007authenticating} addresses the problem of mobile phone users' authentication using keystroke analysis. A feasibility study to demonstrate the ability of neural network classifiers based on users' keystroke and typing dynamics is presented. It is shown that the performance of the technique used can differ due to two reasons: users with large variations in handset interactions, and users who do not use their mobile handset keypad.

\paragraph{}\cite{zahid2009keystroke} demonstrate that keystroke dynamics can be translated to authenticate users. They collected data from $25$ different smart phone users considering six distinguishing keystroke features and demonstrated that these features can be used as a matter of user authentication. In $2014$, \cite{darabseh2014accuracy} studied the influence of four keystroke features and their combinations including key duration, flight time latency, diagraph time latency, and word total time duration. Their computational results on eight users data set confirms that holding time of the key press (F1) is an important feature, among others.

\item Touch Dynamics

\paragraph{}With the increased popularity, usage, and capabilities of touch-screen devices (such as smart-phones, computers, and tablets), users tend to store their personal and sensitive information (such as online banking transactions, PINs, credit card numbers, online transaction credentials, and email communications) on them. Despite its convenience, the stored data is easily attacked by cyber criminals. Therefore, the need for sufficient user authentication schemes to protect both users and companies is increasing. One of the recent approaches for properly authenticating users utilizes touch dynamics and characteristics of the user. Touch dynamic authentication has some advantages, such as that it is an inherent feature of a many smart-phone and computer devices already (\cite{teh2015recognizing}). Furthermore, touch dynamic schemes use the internal sensors of the touch screen device and does not require any extra hardware (\cite{inoue2018taponce}).  
Due to the advantages of touch dynamic schemes mentioned in the literature,
 recent studies of touch dynamics are further described and addressed in this section.

\paragraph{}In the literature, machine learning methods are the main tool for addressing and constructing user authentication schemes based on behavioral biometrics. In 2012, \cite{meng2012touch} proposed a novel user authentication approach based on 21  features related to the touch dynamics of a user such as touch duration and direction. To validate the performance of their method, they used 20 Android phone users and the results show that the neural network machine learning classifier effectively authenticated
users. It is also shown that with the usage of Particle Swarm Optimization (PSO), the average error rate of variations in users' usage patterns was
reduced to $3\%$.

\paragraph{}One year later, \cite{frank2013touchalytics} investigated the usage of touchscreen input as a behavioral biometric for continuous user authentication and developed a touch behavioral authentication scheme called \textit{Touchalytics}. By considering 30 touch dynamic features and using k-nearest neighbour classifier as well as a Gaussian rbf kernel support vector machine on data collected from $41$ users, they show that classifiers achieve robust results with an error rate between $0\%$ and $4\%$. \cite{sae2014multitouch} proposed a user authentication scheme which considered
22 different features of multi-touch behavior gestures. These gestures could be extracted from both hand and figure gestures including pinch, drag, and swipe. Also, a multi-touch gesture classifier was developed.

\paragraph{}A study by \cite{meng2014design} addresses a novel lightweight touch dynamic based user authentication system, considering $8$ features such as the number of touch movements, the average time duration of single-touch, and the average touch pressure. They also maintain the accuracy of classifiers by designing an adaptive mechanism and measuring the performance of classifiers with a cost-based metric. An experiment considering $50$ users using Android phones was conducted and it was shown that the proposed authentication scheme can achieve an average error rate of $2.46\%$. 
\paragraph{}In a more recent article studied by \cite{gong2016forgery}, a new touch-based continuous authentication system, secure against forgery attacks, is proposed. The authors consider some random "\textit{secret}" in a user's touch characteristics that an attacker cannot be aware of even if the user's touch characteristics were already obtained. To illustrate the results, data from $25$ users was collected, and it was shown that the proposed model was able to achieve a smaller equal error rate than previous touch dynamic authentication schemes. \cite{sharma2017user}'s study on user authentication and identification from user interface interactions shows that some features are not particularly useful, such as the difference in coordinates, distance, finger pressure, finger size, and direction of touch. They collected data from $42$ users and found that the \ac{SVM}-based classifier outperformed the other two techniques used and achieved a mean equal error rate of $7\%$ for user authentication. The results also demonstrate that the median accuracy of $93\%$ was achieved for user identification.

\paragraph{}Although machine learning methods are widely used for user authentication schemes, the performance of a classifier may not be stable. A recent study by \cite{meng2018enhancing} addresses this instability performance of the machine learning techniques used for the user authentication and proposes a cost-based intelligent mechanism to choose the least costly mechanism while maintaining the performance of the user authentication scheme. To achieve this goal, they design a lightweight touch dynamics based authentication scheme that includes $9$ gesture-related features such as the the number of touch movements, the average time duration of touch movements, the average speed of touch movement, touch size, and touch pressure. The computational results on data collected from $60$ Android phone users demonstrate that the proposed cost-based authentication scheme achieves a higher and more stable level of authentication accuracy.

\paragraph{}\cite{rehman2017authentication} analyzes the distinctness of gestures in touchscreen Android mobile devices to measure the accuracy, distinctness of user's fingers, and time response of the Android mobile device. The results show that the accuracy of finger gestures can be increased by a user's right index fingers, left index finger, left thumb, and right thumb. In addition, the phone’s position, orientation, screen size, and the dominant hand of the user and the number of hands used can affect the accuracy. 

\paragraph{}Moreover, in another study by \cite{meng2018touchwb}, \textit{TouchWB},
a novel touch gesture-based authentication scheme consisting of $21$ touch features, is introduced. Features of the study are extracted from web browsing gestures and behaviors of the users on smart-phones and can be mainly categorized in four types: single-touch (such as tap), multi-touch (such as zoom, pinch, and rotate), touch-movement (such as swipe up and swipe down), and no-touch. In addition, authors consider $8$ different directions to describe and define touch movements. In order to investigate the touch behavioral deviation between web browsing and freely using the smart-phone, data from $48$ Android phone users divided in two groups was collected. The results demonstrate that the deviation of user's touch behavioral during web browsing is smaller than the scenario of free touches and the combined classifier of PSO and radial basis Function Network (RBFN) adopted from \cite{meng2012touch} could achieve a mean error rate of $2.38\%$.

\paragraph{}The study of \cite{inoue2018taponce} challenges the limit of usability in touch dynamic authentications and addresses a novel user authentication scheme, called \textit{TapOnce}. It is designed to authenticate users with only one tap, while previous works required at least four taps for touch dynamic authentication (\cite{inoue2017one}). Authors discuss that among the three indispensable factors of security, usability, and system efficiency, usability is considered the key factor; their proposed scheme, considering $211$ features, is able to enhance the usability of mobile device authentications. Their experimental results based on $10$ users and $25$ unauthorized users show an average error rate of $3.8\%$ without overfitting.

\end{itemize}

\subsubsection{Combined Authentication}
\paragraph{}As described in previous sections, common user authentication schemes vary from password-based to biometric-based methods. Under such an approach, users are authorized based on only one of the aforementioned methods. However, some recent studies have shown that the combination of two or more above authentication schemes can be utilized to authenticate users and enhance the accuracy of authentication system. Some of the combination methods include face and touch authentication; signature and touch , authentication; PIN and touch authentication; and unlock pattern and movement-based touch authentication. In this section, some of the recent studies in combined authentication methods will be discussed.

\paragraph{}\cite{de2012touch} introduce a two-factor authentication scheme by using both the unlock screen password pattern and the touch dynamics performed to perform that pattern. It investigates the use of touch dynamics to improve the security of the login process and decrease the chance of a shoulder surfing attack. Their results on a lab and a long-term study provided the first proof that the security of password patterns can be increased by considering the touch dynamics. Similarly, \cite{meng2016tmguard} show that after several trials, users perform the same pattern somewhat stably. Due to this behavioral biometric, they develop a new touch movement-based user authentication scheme that authenticate users by combining users' touch movements and Android unlock patterns called \textit{TMGuard}. The authors evaluate the proposed scheme by collecting data from $75$ users, and show that \textit{TMGuard} can enhance the security of Android unlock patterns while maintaining its usability. \cite{meng2016evaluating} evaluates the effect of multi-touch behaviours on the creation of Android unlock patterns. The author designs two user studies with a total of $45$ users and shows that multi-touch movements have an impact on creating unlock patterns and graphical passwords. 

\paragraph{}\cite{pasenchuk2016signtologin} introduced SignToLogin, a two-factor biometric authentication scheme that solves some drawbacks of the previous two-factor authentication approaches. It uses login ID and a push notification to authenticate users in two stages. The push notification is sent to users and asks them to enter their biometric as a second factor. This new push notification replaces the need for text messages which was used in insecure channels. Experimental results show that SignToLogin achieves an equal error rate
of $2\%$.

\paragraph{}Later on, \cite{van2017draw} proposed the \textit{DRAW-A-PIN} user authentication method in which the  touch behavior of a user while drawing a PIN is used to authenticate the user. \textit{DRAW-A-PIN} asks users to draw their PIN instead of typing it on a keypad on touch screens to  analyze user touch behavior. To evaluate the performance of \textit{DRAW-A-PIN}, data from $20$ volunteers over $10$ days and imitation samples from $25$ attackers was collected; the results indicate that in the scenario where the attacker knows the PIN, \textit{DRAW-A-PIN} achieves an equal error rate of $4.84\%$. Moreover, in the scenario where the attacker knows how to reproduce the exact same drawn PIN, \textit{DRAW-A-PIN} rejects attackers at a rate of $85\%$.

\paragraph{}The study of \cite{zheng2014you} found that a user's tapping signature combined with their PINs can reduce the chance of any possible attacks and increases the accuracy of user authentication. In this manner, based on collected data from over $80$ users and exploiting the combination of four features, they show that the proposed authentication scheme has high accuracy with average error rate of $3.65\%$. In a similar study, \cite{sun2014touchin} introduces a two-factor authentication approach using both the geometric properties of a user's drawn curves and its behavioral and physiological characteristics. The proposed method, TouchIn, achieved high security, efficiency, and usability. the next year, \cite{chen2015your} introduced RhyAuth, a two-factor authentication scheme based on the performance of users on a sequence of rhythmic taps and/or slides on touch screen devices. RhyAuth uses the user-chosen rhythmic and touch biometrics while inputting that rhythm. Using collected data from $32$ users, the authors show that their proposed method is highly secure against attackers. 

\paragraph{}Researchers from IBM (\cite{trewin2012biometric}) conducted research to examine user time, effort, error, and task disruption on password entry and three biometric authentication schemes, i.e., voice, face, and gesture. Their results in a laboratory study show that face and voice authentication schemes were faster than password entry, while voice was less usable than the other three.

\paragraph{}\cite{shahzad2017behavior} propose a novel authentication scheme, called \textit{BEAT}, that considers behavioral biometrics and certain actions, i.e., a gesture or a signature. A gesture can be an interaction of a user with the touch screen while a signature is a unique handwritten depiction of the user's name. This authentication method helps in preventing attacks such as shoulder surfing while the attackers cannot reproduce the touch behavior of the user. To perform the computational results, data from $25$ users are collected, and it is shown that \textit{BEAT} with seven types of features (i.e., velocity magnitude, device acceleration, stroke time, inter-stroke time, stroke displacement magnitude, stroke displacement direction, and velocity direction) achieves an average equal error rate of $0.5\%$ with three gestures and $0.52\%$ with a single signature.

\paragraph{}In a similar study, \cite{buriro2015touchstroke} proposes a new combined authentication scheme for touch-typing behavior of the user called "\textit{Touchstroke}". They specifically use data from how the user holds their phone and how the $4$-digit text-independent PIN is entered. The holding scheme is determined by $7$ built-in sensors including orientation, gravity and gyroscope. Random Forest and BayesNET are used to test the data collected from $12$ users. The results show highly accurate user authentication. 

\paragraph{}Recently, \cite{buschek2018researchime} propose a filtering concept to log typing biometrics of $349$ users without recording a text's content. They released an Android keyboard app to implement the concept they proposed. With this free text entry study, in three weeks, authors presented the first analyses of both keyboard use and typing biometrics based on unfiltered real world evidence. Some of the biometric features considered are speed, posture, apps, auto correction, and word suggestions. They finally concluded that free and unconstrained studying of typing biometrics of users on a daily basis can produce different results than inside lab experiments.

\paragraph{}In 2014, \cite{kang2014two} proposed a two-factor identification scheme based on username/password and face recognition authentication schemes. Following the same idea, the study of \cite{smith2016continuous} presented a novel stacked classifier approach to implement a user authentication framework considering both face and touch authentication schemes. Authors use a public data set of face and touch-gesture modalities for $50$ users, and the results show that the proposed multi model framework reduces the false acceptance risk and achieves an equal error rate of $3.77\%$ for a single sample. 

\paragraph{}\cite{song2017multi} conducted
research on authenticating users based on both hand geometry and behavioral characteristics. Specifically, users are asked to perform specially designed multi-touch gestures (\textit{TEST} gestures which require users to stretch their fingers and put them together) with one swipe on multi-touch screens devices. To evaluate the results, authors analyze data collected from $161$ users and show that the proposed method achieves an average error rate of $5.84\%$ with $5$ training samples and $1.88\%$ with enough training samples.

\paragraph{}This year, \cite{buriro2018dialerauth} introduced a user authentication scheme that uses both user tapping/touching behavior and hand micro-movements while entering a text-independent 10-digit number, called \textit{DIALERAUTH}. Experimental results on data collected from $97$ users prove the resilience of \textit{DIALERAUTH} against random and mimic attacks while achieving usability and acceptance by a mean score of $73.29$. 

\paragraph{} A study by \cite{neal2017using} addresses a user authentication scheme using associative classification. As a matter of behavioral biometrics, they provide a performance analysis of applications, Bluetooth, and Wi-Fi usage data and collect data from $189$ users. Named association rules are extracted and combined as a feature extraction technique to demonstrate the effectiveness of associative classification. Results indicate the accuracy of applications and Bluetooth traffic with an accuracy rate of $91\%$.

\paragraph{}Regarding a more novel topic, implicit authentication, \cite{shen2018performance} addressed a continuous and implicit user authentication scheme using motion sensors inside smart-phones to monitor users' daily activities. They establish a data set collected from $10$ users while considering five typical human daily activities and five phone placements. Authors present the average error rate of their proposed method and show that their proposed approach is feasible only in some real scenarios. Another article to address is presented by \cite{akhtar2017multimodal}, which presents an unconstrained and implicit smart-phone multimodel biometric system. Micro-movements of the smart-phone, movements of the user’s finger during typing on the touchscreen, and the user’s face features are the phone movements, touch strokes,  and face patterns that the article uses to authenticate users. Computational results on the data collected from $95$ users show high accuracy while increasing the security and usability.  

\section{Conclusion}

\justify{
In this section, we present some of the key findings of our survey. During this survey, we noticed that supervised learning techniques have been used more frequently than unsupervised methods. To be more specific, the most commonly used fraud detection techniques are \ac{LR}, \ac{ANN}, \ac{DT}, \ac{SVM} and \ac{NB}. As a matter of fact, according to the study by \cite{albashrawi2016detecting}, among the most applied methods in a period ranging from $2004$ to $2015$, \ac{LR} seems to be the leading technique in detecting financial fraud with $13\%$ usage frequency, followed by both \ac{ANN} and \ac{DT}, with $11\%$. \ac{SVM} and \ac{NB} are represented by $9\%$ and $6\%$ frequency, respectively. Moreover, these techniques have been used alone or combined with an ensemble technique to build strong detection classifiers.

Though the frequency of using supervised methods could be due to the fraud context and the availability of labeled data, it could also be also based on the performance of these methods. In other words, supervised learning techniques might be better-performing tools than the unsupervised ones in detecting financial frauds. Another reason for under-employment of unsupervised learning methods could be the challenge of coming up with a reasonable objective measure of the algorithm’s accuracy, since there is no set standard (like a target output or a label). Therefore, choosing an appropriate measure to assess the similarity of the instances or to compare different structures in the data is always challenging in this learning framework. This a limitation of unsupervised methods in general and not specific to the problem of fraud detection. Therefore, in the availability of annotated transactional data, it is recommended to use a supervised method, due to its advantages.

We also discussed some of the most recent and relevant articles in the field of user authentication. In general, the articles in this area can be categorized as one of the following:}}
\end{flushleft}

\begin{enumerate}
\item Password-based authentication approaches, which is considered for passwords, Personal Identification Numbers (PINs), or Graphical passwords based authentication schemes. This category of user authentication approaches are usable and convenient, however, they are highly insecure and users are more likely to choose simple passwords due to long term memory limitations. In addition, these approaches can be stolen via shoulder surfing and can be recovered via side channel attacks. Therefore, researches suggest using a new approach based on human body measurements or actions, called biometric-based approaches.
\item Biometric-based authentication approaches, which can be categorized in two subsections:
physiological authentication, and behavioral authentication approaches. The former addresses authentication schemes based on human body measurements and its physical characteristics such as the
face, finger, iris, voice, or retina. Similar to password-based authentication approaches, physiological approaches have some limitations and cannot be efficient for authentication. They mostly suffer from requiring additional and special hardware to scan and/or recognize the physiological features of users. Also, it is limited to one authentication at the beginning of a session. Finally, it can be highly dependent on environmental factors, such as different viewing angles, poor illumination, and background noises. It is important to mention that some of the devices used to detect these features such as retina or iris are very expensive. The latter biometric-based authentication scheme, called behavioral authentication, is allocated to the behavioral features of users such as keystroke dynamics or touch dynamics of users. This sub category of biometric-based authentication schemes is considered the most recent and successful approach in the literature. 
\item Combined authentication approaches, which are considered one of the most recent approaches in user authentication schemes. These approaches consider more than one of the aforementioned categories to increase the accuracy of the system while maintaining its usability.
\end{enumerate}
\begin{flushleft}
\justify{

During our survey, we came across a study \citep{khan2014comparative} that evaluated several authentications schemes on four independently collected data sets from over $300$ participants. They evaluated different schemes in various categories including but not limited to accuracy, training time, detection delay, processing and memory complexity for feature extraction, and training and classification operations. To be more specific, they compared gait pattern, browsing behavior, keystroke dynamics, and different touch behavior-based authentication schemes in terms of the aforementioned criteria.

Their evaluations showed that in addition to adequate data availability for training and classification, touch behaviour-based authentication schemes outperform other schemes in terms of accuracy and detection delay. Also, another interesting finding is that the authentication schemes that rely on input events (e.g. touch- and keystroke-based schemes) at the device level require root privileges on the device in order to collect data. On the other hand, input event data can be collected by individual apps without any additional permission, which opens the door for IA protection at the app level instead of at the device level. It is also noticed that keystroke features do not appear to be distinctive even in scenarios with few subjects, and this suggests that it would be wise to pair these features with other, behavior-derived attributes when creating authentication schemes.

Finally, their findings showed that having users perform a repeated task in a lab setting to generate data leads to unrealistic evaluations. When these type of authentication schemes are then applied in real-world settings, the assumptions made in the lab may prove false and the scheme’s performance will suffer accordingly. It has been demonstrated in their study that on real-world data sets, many existing touch-based authentication schemes have significantly higher equal error rate than reported from lab experiments.

Through our study here, we also experimented on a CMU keystroke dynamics data set to detect impersonations in a password login system. We tested a variety of supervised learning methods, and used the equal error rate to evaluate the performance of each learning algorithm. The experiments showed that the models with less parameters yielded lower equal error rates compared to that of a deep learning-based algorithm. The main reason behind this under-performance of the \ac{NN}-based model was perhaps the small number of samples. Deep learning is designed to offer a very high degree of freedom which enables learning complex patterns. This means that deep networks are composed of too many parameters and variables to be tuned. However, in order to set the proper values for these multiple variables (weights and biases), a large amount of training data is usually required. In terms of error rate, the random forests model seems to perform better than any of the other methods.

In order to acquire better results, we pre-processed the data to filter out noises of each subject. As our experiments show, this filtering leads to lower error rates compared to the situation where models were applied on noisy data. These results indicate the importance of noise reduction to remove the samples that are not indicative of normal behavior of the subject (i.e. distraction, inebriated, etc.). Thus, we determine a mean vector for each subject in our data set and drop all training data points greater than 3 standard deviations from the mean vector. This is a commonly used pre-processing step to remove noise from data. As demonstrated by the results, this noise reduction step leaded to a decrease in the equal error rate in all experiments with different algorithms. Finally, utilizing this data pre-processing technique combined with the \ac{RF} model, we reached an average equal error rate of $~3.5\%$ over all subjects over $30$ runs. It is worth pointing out that this is also consistent with what is reported in the literature confirming the superiority of \ac{RF} compared to other alternatives (see for example \cite{kumar2018continuous, lee2016touch, feng2012continuous, watanabe2017long,centeno2018mobile}).}
\end{flushleft}

\section{Future Directions
}
\begin{flushleft}
\justify{
Recent research suggests that the inclusion of contextual information--- collected along side of the behavioral profiles---leads to more stable and accurate profile-based intrusion detection systems \citep{yampolskiy2008generation,kim2006baseline,sommer2003enhancing}. However, obtaining biometric data in sufficient quantity is always challenging, due to the privacy of individuals providing the data. One solution to this problem could be generating large quantities of high quality synthetic yet realistic biometric data, which would lead to the development of more robust anomaly detection models. The main focus and goal for such a simulation would be to yield a completely self-sufficient data set with the goal of having similar statistical properties as the original data set. To yield such results, the simulator must go through several steps. The concepts that are behind the model are based on statistical analysis of large batches of real data.

This data should be generated in such a way that it includes more modalities on the behavior of the users. For example, the data might include a set of features on combination of movement, orientation, touch, and gesture behaviors of users on their smartphones. The expectation is that using more modalities has
the potential to reduce error rates of state-of-the-art authentication models that only use touch or phone movement features in isolation. Also, including a more comprehensive set of features enables conducting a feature analysis, which help determine the features that are (more) responsible for a particular decision. Moreover, generating a simulated set of data creates the opportunity to investigate the power of deep learning models when dealing with large-scale context-informed behavioral data. 

}
\end{flushleft}

\bibliographystyle{IEEEtran}
\bibliography{RefUserAuth}

\end{document}